\documentclass{article}[12pt]
\usepackage[a4paper, total={6in, 8in}]{geometry}
\usepackage[utf8]{inputenc}
\usepackage{amsmath,amssymb,dsfont,amsfonts,amsthm,amscd}
\usepackage{tabularx}
\usepackage{graphicx}
\usepackage{color}
\usepackage{array}
\usepackage{url}
\usepackage{subcaption}
\usepackage{bm}
\usepackage{authblk}
\usepackage{natbib}
\usepackage{dsfont}
\usepackage{float}
\usepackage{adjustbox}
\usepackage{booktabs}
\usepackage{multicol}

\theoremstyle{plain}

\theoremstyle{definition}

\theoremstyle{remark}

\title{One-for-many Counterfactual Explanations by Column Generation}
\author[1]{Andrea Lodi}
\author[2]{Jasone Ram\'{\i}rez-Ayerbe}

\affil[1]{Cornell Tech, Cornell University}
\affil[2]{Instituto de Matem\'aticas de la Universidad de Sevilla\\}
\affil[ ]{ \tt{andrea.lodi@cornell.edu}}
\affil[ ]{\tt {mrayerbe@us.es}}

\date{}

\begin{document}

\maketitle

\begin{abstract}
In this paper, we consider the problem of generating a set of counterfactual explanations for a group of instances, with the one-for-many allocation rule, where one explanation is allocated to a subgroup of the instances. For the first time, we solve the problem of minimizing the number of explanations needed to explain all the instances, while considering sparsity by limiting the number of features allowed to be changed collectively in each explanation. A novel column generation framework is developed to efficiently search for the explanations. Our framework can be applied to any black-box classifier, like neural networks. Compared with a simple adaptation of a mixed-integer programming formulation from the literature, the column generation framework dominates in terms of scalability, computational performance and quality of the solutions.

\end{abstract}

\section{Introduction}


In recent years, machine learning algorithms have been used in high-stakes decision-making settings, such as healthcare, loan approval, or parole decisions \citep{baesensMS03,zeng2022uncovering,zengJRSC17}. Consequently, there is a growing interest and necessity in their explainability and interpretability \citep{du2019techniques,jung2020,molnar2020interpretable,rudin2022interpretable,zhang2019should}. Once a supervised classification model has been trained, one may be interested in knowing the changes needed to be made in the features of an instance to change the prediction made by the classifier. These changes are the so-called counterfactual explanations \citep{martensMISQ14,wachter2017counterfactual}. There is a growing literature on the development of algorithms to generate counterfactual explanations, see \cite{artelt2019computation,guidotti2022counterfactual,karimi2022survey,sokol2019counterfactual,stepin2021survey,verma2022counterfactual} for recent surveys on Counterfactual Analysis. Nevertheless, they mainly focus on the single-instance, single-counterfactual case, where for one specific instance, a single counterfactual is provided \citep{wachter2017counterfactual,parmentier2021optimal}. Other works consider the case where for one instance, a set of diverse explanations is sought, to give the individual more flexibility to change their class \citep{Mothilal_2020}. The counterfactual analysis is seen as a local explainability tool. 


If we adopt a stakeholder perspective, there are multiple situations where a global explanation is sought. First, stakeholders may want, before deploying the classification model, to be sure that there are no biases in the model \citep{rawal2020beyond}. By computing global explanations for specific subgroups, maybe defined by some sensitive features, unfairness can be detected. Second, for similar instances, instead of assigning to each of them a specific explanation, it has been argued that practitioners and users preferred to be given one single explanation assigned to the group of similar instances \citep{warren2023explaining}. Having one explanation for a group helps find patterns in the explanations \citep{keane2020good} and ensures that similar instances have similar counterfactuals \citep{CARRIZOSAGroup2024}. Third, stakeholders may be interested in detecting the key features that need to change, globally, for all the instances to change their prediction. Many academic discussions \citep{CARRIZOSAGroup2024, miller2019explanation, verma2022counterfactual} emphasize the value of sparser counterfactuals, as they can be more easily interpreted due to fewer features changed. The same sparsity that can be found in individual explanations should be sought in global explanations. Last, stakeholders may be interested in the search for just a few counterfactual instances for a group to be seen as benchmarks of the class \citep{CARRIZOSAGroup2024}.
In this paper, we focus on the two latter tasks. For a group of instances, that have been classified in the negative class by a given classifier, we seek the minimum number of counterfactual instances that cover the whole group. Thus, each counterfactual instance is assigned to a subgroup of the original examples. Moreover, we seek to detect the key features that each subgroup of instances has to change to flip the classifier's decision. We have two levels of interpretability: the detection of both the benchmarks and the key features.

\section{Related work}

As stated before, most of the work on counterfactual explanation focuses on the single-instance, single-counterfactual scenario and there is little literature on the global use of counterfactuals. However, when a stakeholder's perspective is taken, four different situations can appear, as described in \citep{CARRIZOSAGroup2024}. These situations depend on the allocation rules between the original instances and their corresponding counterfactuals. The commonly addressed scenario is when each instance has exactly one counterfactual and vice versa, which is called the one-for-one allocation rule. The second allocation rule, the many-for-one rule, seeks several counterfactuals for one single instance, which is usually used to have diversity in the explanations. In the third and fourth allocation rules, the one-for-all and the one-for-many rules, one counterfactual explanation is allocated to either all or a subgroup of the instances. 

One way to obtain global explanations is to use the one-for-one allocation rule, but impose some linking constraints to the counterfactuals related to perturbing the same features. In doing so, even though one still obtains one counterfactual explanation for each single instance, the key perturbed features are shared for all of them. This is the case studied in \citep{CarrizosaESWA24}, where counterfactual explanations are sought for a group of individuals, and the total number of features perturbed is minimized. 

The same idea is followed in the Group Counterfactual Explanations by \citep{warren2023explaining}. The authors in that paper focus on generating one global explanation for a specific group of instances. It is important to point out that in their case, sharing an explanation means that the altered features and values to obtain a positive counterfactual are the same for all original instances, but the non-altered features may differ for each of the instances, resulting in one unique counterfactual instance for each of the individuals. Thus, their work falls into the single-instance single-counterfactual scenario with linking constraints. Moreover, they focus on pre-existing subgroups, so their method does not allow the detection of them. 

The work of \citet{ley2023globe} does consider the case where subgroups may be detected. In that paper, a subgroup of individuals share a global explanation if they move in the same translation direction, but where the translation direction is multiplied by an example-dependant scalar. Still, each counterfactual instance ends up having its assigned counterfactual. In the case where the subgroups are given, their method offers an efficient runtime in the order of tens of seconds. 

In other related work, \citet{rawal2020beyond} aim to provide recourse summaries for stakeholders to analyze whether a classifier may have some bias or not. The method does not calculate counterfactual instances for a group of individuals, instead, it calculates rules explaining feature-difference within different subgroups. It does not only summarize the counterfactual explanations but also the original individuals. In the end, the method aims to solve a different problem than what we are interested in. 

Finally, a one-for-many allocation model is presented and illustrated in \citep{CARRIZOSAGroup2024}. For a group of instances, a predetermined number of counterfactual instances is calculated. The set minimizes the sum of the Euclidean distance between each original example and its assigned counterfactual and obtains the positive examples for the group, that can be seen as benchmarks for the positive class. To solve this problem, \citet{CARRIZOSAGroup2024} use an iterative clustering algorithm, similar to Lloyd's algorithm, and they only apply it to a logistic regression classifier. As far as we know, this is the only work where the one-for-many allocation rule is considered, i.e., when only a few counterfactual explanations are obtained for a group of negative examples. However, there is no sparsity included in \citep{CARRIZOSAGroup2024}. Each negative example is associated with its closest counterfactual explanations, but the features that need to be perturbed in each of the subgroups are not detected.

\section{Problem Statement/Contribution}

In this paper, we consider the problem of generating collective counterfactual explanations for groups of negative-classified examples referred to in \citep{CARRIZOSAGroup2024} as the one-for-many allocation rule. For the first time, we comprehensively take into account all components of the problem, namely, 
\begin{itemize}
    \item We minimize the number of positive examples explaining all the negative ones. Doing so, we significantly increase \emph{interpretability}, because having fewer counterfactual explanations significantly contributes characterizing the set used to cover all the examples.
	\item We directly consider \emph{sparsity} by limiting the number of features that can be changed in any example and linking the features that are changed in any specific group, i.e., we implicitly detect subgroups of the population that share the same counterfactual explanation. Doing so, we again ease the use of the counterfactuals in the contexts in which interpretability is a requirement.
\end{itemize}
This is achieved by developing a sophisticated mixed-integer programming formulation based on a novel column generation framework. We compare this formulation with a simple adaptation of the mixed-integer programming formulation by \citep{CARRIZOSAGroup2024} and we computationally show the significant advantages in terms of scalability, computational performance and quality of the solutions. The framework incorporates as a black-box any classifier that can be formulated (after training) as a mixed-integer programming problem, for example, neural networks or logistic regression.

Remarkably, by construction, our methodology obtains 100\% coverage of all examples and 100\% precision in terms of ensuring that the changes made in the individuals change the classifier's decision. This is a significant advantage with respect to the literature discussed above.

\section{Problem definition}

Let us consider a probabilistic binary classification problem on a set $\mathcal{X} \subset \{0,1\}^{\bar T}$ with classes $\{-1,+1\}.$ Suppose we have a probabilistic classifier $P: \mathcal{X} \rightarrow [0,1]$, where $P(\bm{x})$ is the probability of belonging to the positive class, i.e., $+1$. We will suppose that all features $x_j$ are binary. In the case where there are numerical or categorical features, discretization followed by one-hot encoding will be considered. 

Let $\{\bm{x}^0_1,\dots,\bm{x}^0_{|S|}\}$ be a set of negative instances or examples. We want to obtain the minimum positive explanations that cover all of them. We want global explanations for the whole group and we want to detect the key features to flip the classifier's decision for each of the groups.



We want to use a small set of features collectively in each explanation. So, for the set of instances assigned to one counterfactual, the perturbed features will be the same, thus resulting in a much more interpretable rule. We will bound the number of features changed in each cluster by a given parameter $T_{\text{max}}$, while our objective will be to minimize the number of total explanations used for the whole group.

The above problem can be formulated through mixed-integer linear programming techniques. We will use the notation in Table \ref{tab:notation}.

\begin{table}[!ht]
    \centering
    \caption{Notation}
    \label{tab:notation}
    \begin{tabular}{p{0.1\linewidth} p{0.8\linewidth}}
        \multicolumn{2}{l}{\bf{Parameters}}
         \\
        $T$ & set of features \\
        $\bar{T}$ & expanded set of features \\
        $T_l$ & set of indices of features in $\bar{T}$ corresponding to original feature $l, l \in T$ \\
        $T_c$ & set of categorical or continuous features \\
        $S$ & set of negative examples, $i \in S$ \\
        $K$ & upper bound on the number of positive counterfactuals needed \\
        $v_{ih}^0$ & value of feature $h\in T$ of original example $\bm{x}_i^0 \in \{0,1\}^{|\bar{T}|}, i\in S$ \\
        $T_{\text{max}}$ & maximum number of features to be perturbed \\ 
        \addlinespace 
        \multicolumn{2}{l}{\bf{Decision variables}}\\
        $y^k$ & binary variable indicating if counterfactual $k$ is used, $k=1,\dots,K$ \\
        $v_h^{k}$ & value of feature $h\in \bar{T}$ of counterfactual explanation $\bm{x}^k$ , i.e., $\bm{x}^k=(v_1^k, \dots, v_T^k)$, $k=1,\dots,K$ \\
        $a_i^k$ & binary variable indicating if instance $i$ is explained by the counterfactual $k$, $i\in S$, $k=1,\dots,K$ \\
        $d_h^k$ & binary variable indicating if feature $h$ is used in the explanation $k$, $h\in \bar{T}$, $k=1,\dots, K$ \\
        $f_l^k$ & binary variable indicating if original feature $l$ changes in explanation $k$, $l\in T$, $k=1,\dots,K$ \\
        $\xi_{ih}^k$ & binary variable indicating whether feature $h\in T$ is different between instance $i\in S$ and counterfactual $k$, $k=1,\dots,K$.
    \end{tabular}
\end{table}
Notice that we have two sets of features. The original set of features $T$ reflects the features of the dataset, which can have categorical and/or numerical features, whereas the set $\bar{T}$ represents the expanded set of features, after the one-hot encoding, and thus all features are binary. 

\subsection{Baseline formulation}


Following \citep{CarrizosaESWA24}, the global counterfactual explanation problem can be defined as a mixed-integer program (MIP), where the number of explanations is minimized while ensuring that each example is compatibly covered. In this context, compatibility means that every negative example assigned to the same counterfactual explanation perturbs the same set of features, and at most $T_{max}$ features. 



The MIP formulation reads as follows:

\begin{align}
	\min \quad &  \sum_{k=1}^K y^k\label{eq:milp_obj} \\
	\text{s.t.} \quad &P(\bm{x}^k) \geq \tau, \quad \forall k=1,\dots,K\label{eq:milp_cons0} \\
	& d_h^k \geq \xi_{ih}^k a_i^k \quad i \in S, h \in \bar{T}, k=1,\dots,K \label{eq:milp_cons1}\\
&|T_l| f_l^k \geq \sum_{h\in T_l} d_h^k \quad l \in T, 
 k=1,\dots,K \label{eq:milp_cons2}\\
&\sum_{l \in T} f_l^k \leq T_{\text{max}}y^k, \quad k=1,\dots,K\label{eq:milp_cons3}\\
&  \sum_{h \in T_l} v_h^{k}=1 \quad l\in T_c, k=1,\dots,K \label{eq:milp_cons4}\\
	&\sum_{k\in K} a_i^k\geq 1 \label{eq:milp_cons5}\\
 & a_i^k \leq y^k \quad \forall i \in S, k=1,\dots,K \label{eq:milp_cons6}\\
	&v_{ih}^0-v_h^k \leq \xi_{ih}^k \quad \forall i,h,k \label{eq:milp_cons7} \\
	&v_{ih}^0-v_h^k \geq -\xi_{ih}^k \quad \forall i,h,k. \label{eq:milp_cons8}
\end{align}

With the objective \eqref{eq:milp_obj} the number of counterfactual explanations used is minimized. Constraints \eqref{eq:milp_cons0} ensure that the counterfactual explanations are classified as positive. Constraints \eqref{eq:milp_cons1} define the expanded features used in each explanation, while constraints \eqref{eq:milp_cons2} define the original features used. Constraints \eqref{eq:milp_cons3} bound the size of the explanations. Constraints \eqref{eq:milp_cons4} preserve the coherence of the categorical or numerical features where one-hot encoding has been used. Constraint \eqref{eq:milp_cons5} imposes that each negative example is associated with at least one counterfactual explanation. Constraints \eqref{eq:milp_cons6} impose that if an explanation is not used, then no explanation can be associated with it. Finally, constraints \eqref{eq:milp_cons7} and \eqref{eq:milp_cons8} model the absolute value of the change. 

To linearize constraint \eqref{eq:milp_cons1} a new binary variable $\gamma_{ih}^k \in \{0,1\}$ for each $i, h, k$ is introduced and the following constraints are added to the formulation: 

\begin{gather}
	d_h^k \geq \gamma_{ih}^k\\
	\gamma_{ih}^k \leq \xi_{ih}^k\\
	\gamma_{ih}^k \leq a_{i}^k \\
	\gamma_{ih}^k \geq \xi_{ih}^k+a_i^k-1. \label{eq:milp_consf}
\end{gather}

The linearization of formulation \eqref{eq:milp_obj}-\eqref{eq:milp_cons8} above is only practical to solve for small datasets. In the next section, we will propose an alternative formulation based on column generation \citep{barnhart1998branch}.

\subsection{Column generation formulation}

Let $\mathcal{K}$ be the set of all possible counterfactual explanations. This set is clearly exponential, i.e., unpractical for any interesting size, and the key idea of column generation (CG) is to start from a small set of counterfactual explanations and generate additional ones on the fly until we have enough to prove optimality. This is obtained by exploiting linear programming (LP) duality as discussed in the following.

We will start with a small subset $\mathcal{K}^{*} \subset \mathcal{K}$ of possible counterfactual explanations. With a slight change of notation,\footnote{We move the superscript $k$ in the previous formulation to be a subscript to indicate the difference of the formuations.} let $a_{ik}$ be 1 if negative example $i \in S$ can be explained by counterfactual $k \in \mathcal{K}^{*}$,  i.e., less than $T_{\text{max}}$ features need to change. Then, to solve the restricted problem, the following MIP is proposed:

\begin{align}
\min\quad &   \sum_{j\in \mathcal{K}^{*}}  y_k \label{eq:rmp_obj}\\
\text{s.t} & \sum_{\substack{k\in K^{*}}} a_{ik} y_k \geq 1, \quad i \in S \label{eq:rmp_cons}\\
&  y \in \{0,1\}^{|\mathcal{K}^{*}|}. \label{eq:rmp_cons2}
\end{align}

Notice how, in MIP \eqref{eq:milp_obj}-\eqref{eq:milp_cons8}, $a_i^k$ were variables of the problem that now correspond to $a_{ik}$, i.e., data. We will call \eqref{eq:rmp_obj}-\eqref{eq:rmp_cons2} the restricted master problem (RMP), while the full master problem is defined by replacing $\mathcal{K}^{*}$ with  $\mathcal{K}$ in \eqref{eq:rmp_cons}. The validity of the full master problem formulation follows directly from selecting the smallest number of positive counterfactual in $\mathcal{K}$ that explain all negative examples in $S$, and clearly depends on the way the counterfactuals are built, which is discussed in the following.

To apply the column generation framework, we will solve the LP relaxation of the master problem, obtaining the optimal solution $\bar y$ and its corresponding dual solution\footnote{We assume to have constructed $\mathcal{K}^{*}$ such that the LP relaxation of \eqref{eq:rmp_obj}-\eqref{eq:rmp_cons2} is feasible, see the discussion later in the section.} $\bar w$. Clearly, $\bar y$ is a feasible solution of the (LP relaxation of the) full master problem but we need to check if it is also optimal. Because of duality, this boils down to check if $\bar w$ is feasible for 
the dual of the (full) master problem

\begin{align}
    \max \quad &\sum_{i\in S} z_i\\
    \text{s.t} \quad & \sum_{i\in S} a_{ik} w_i \leq 1, \quad k\in \mathcal{K} \label{eq:dual_cons}\\
    &z_i \geq 0, \quad i \in S,
\end{align}
i.e., find a counterfactual $\bar k \in \mathcal{K} \setminus \mathcal{K}^{*}$ such that 
\begin{equation}
    \label{eq:violation}
    \sum_{i\in S} a_{i{\bar k}} \bar w_i > 1,
\end{equation}
if any exists. Such a $\bar k$ then 
violates constraint \eqref{eq:dual_cons}. 

To find such a counterfactual $\bar k$, we need to guarantee (i) compatibility, i.e., the fact that the examples explained by $\bar k$ have to change at most the same $T_{\text{max}}$ features and (ii) $\bar k$ is classified as positive by the given classifier. This can be cast as a MIP, where condition \eqref{eq:violation} can be used in the objective function and binary variable $z_i$ indicates whether counterfactual explanation $\bar k$ explains negative example $i \in S$ or not. This is the so-called pricing problem

\begin{align}
\max\quad &   \sum_{i \in S} \bar w_i z_i \label{eq:pp_obj} \\
\text{s.t} \quad & P(v_1, \dots, v_T) \geq \tau
\label{eq:pp_cons1}\\
&d_h \geq |v_h^i-v_h|z_i \quad i \in S, h \in \bar{T} \label{eq:pp_cons2}\\
&|T_l| f_l \geq \sum_{h\in T_l} d_h \quad l \in T \label{eq:pp_cons3}\\
&\sum_{l \in T} f_l \leq T_{\text{max}} \label{eq:pp_cons4}\\
&  \sum_{h \in T_l} v_h=1 \quad l\in T_c. \label{eq:pp_cons5} 
\end{align}

Constraint \eqref{eq:pp_cons1} ensures that the explanation that we are building is classified as positive. Constraints \eqref{eq:pp_cons2} define the extended features used in the explanation, whereas constraints \eqref{eq:pp_cons3} relate it to the original features. The number of features changed for all the examples covered by $\bar k$ is bounded with constraint \eqref{eq:pp_cons4}. Finally, constraints \eqref{eq:pp_cons5} ensure the correct one-hot encoding of the categorical and numerical features. 

If the optimal solution value to the pricing problem \eqref{eq:pp_obj}-\eqref{eq:pp_cons5} is strictly bigger than 1,  then condition \eqref{eq:violation} is satisfied and $\bar k$ can be constructed as $\bar k = (v_1^{*}, \dots, \bar v_T^{*})$, where $v^*$ is the optimal solution vector of the pricing problem. Once added to the RMP, counterfactual example $\bar k$ could potentially improve the objective function value and the algorithm iterates until the optimal solution value of the pricing problem is less or equal than 1. 
In that case, the optimal solution of the restricted problem coincides with the relaxation of the master problem, while, likely, only a much smaller number of counterfactual explanations have been added with respect to the full set $\mathcal{K}$. Moreover, if the optimal solution of the RMP relaxation is found to be integral, it is then an optimal solution to the original problem. Even when the solution of the RMP relaxation is fractional, the ceiling of the solution serves as a lower bound to the optimal value. The MIP optimal solution is also certified if the lower bound coincides with the solution when solving the final RMP and imposing variables integrality. If neither of the two conditions hold, then to guarantee that formulation \eqref{eq:rmp_obj}-\eqref{eq:rmp_cons2} is solved to optimality one needs to branch and continue with column generation at any node of the tree, i.e., implementing the branch-and-price scheme \citep{barnhart1998branch}.

Finally, as observed in the footnote, an initial feasible solution has to be calculated for the first RMP. It is easy to see that one can compute a counterfactual for each example and start with the identity matrix, meaning associating each example with its counterfactual. The initial solution can be refined by examining whether any counterfactuals are identical, allowing for the grouping of instances in that case. 

\subsection{Algorithm refinement}

To speed up the overall solution process, several strategies are implemented. Using the fact that one counterfactual explanation may have different assignments (with different feature changes) allocated and likewise, a specific assignment may have different explanations possible, once the pricing problem has been solved, two additional MIPs are solved. Fixing the counterfactual explanation, we solve the corresponding MIP and get (if it exists) a new assignment. Analogously, we get a new counterfactual explanation by fixing the assignment obtained. Moreover, when solving the pricing, the solver may find several columns violating constraint \eqref{eq:dual_cons}. It is possible to recover all these solutions and add them in the same iteration. Last, even if a counterfactual explanation is found but it does not violate \eqref{eq:dual_cons}, the algorithm may keep it in a pool, to add it later if, once the new dual variables are calculated, ends up violating the constraint. 

\subsection{Mixed-integer formulations for classification}
\label{sec:classification}

Constraints \eqref{eq:milp_cons0} in the MIP and \eqref{eq:pp_cons1} in the pricing problem are used to guarantee that the counterfactual explanations built by the optimization models are indeed classified correctly as positive examples. Following the seminal work by \cite{Fischetti2018}, this is obtained by including \emph{pre-trained} ML models (both Neural Networks and logistic regression, see below) within the optimization formulations. In other words, the ML models are trained offline and then are incorporated within the mixed-integer linear programming formulations by fixing some of their parameters (e.g., the weights of the Neural Networks), while others are kept variable (e.g., the ReLUs in Neural Networks). Several implementations are currently available of this approach, see, e.g., \citet{JANOS}, \citet{Gurobi}.

\section{Experimental results}

\textbf{Datasets:} We employ 3 datasets to asses our method.  Table \ref{tab:datasets} reports their name, number of instances and number of original and extended features. The first dataset is the \textbf{COMPAS} dataset, collected by ProPublica \citep{angwin2019machine} as part of their analysis on recidivism decisions in the United States. Following \cite{karimi2020model, Mothilal_2020, parmentier2021optimal}, we consider 5 features that, after performing discretization followed by one-hot encoding, result in 15 extended features. The second considered dataset is the \textbf{German credit} dataset \citep{Dua2019} that captures financial and demographic information of loan applicants. Finally, our third dataset is the \textbf{Students performance} dataset \citep{misc_student_performance_320} that includes student grades, demographic, social and school related features and aims at predicting if a student will pass a course or not.

\begin{table}[H]
    \centering
    \caption{Summary of the datasets used for the experiments}\label{tab:datasets}
    \resizebox{0.9\columnwidth}{!}{%
    \begin{tabular}{lrrr}
        \toprule
		& \textbf{No. of }& \multicolumn{2}{c}{\textbf{No. of Features}}\\
        \textbf{Dataset} & \textbf{Instances} & \textbf{Original} & \textbf{Extended} \\
        \midrule
        COMPAS & 6172 & 5 & 15 \\
        German Credit & 1000 & 20 & 59 \\
        Students performance & 395 & 30 & 85 \\
        \bottomrule
    \end{tabular}}
\end{table}

\textbf{Baseline:} We compare the efficacy of our column generation framework with MIP formulation \eqref{eq:milp_obj}-\eqref{eq:milp_cons8}. In fact, there is no prior work that solves the problem of minimizing the number of counterfactual explanations needed to cover all the examples. Thus, besides demonstrating a viable approach for solving such a problem, the goal of our comparison is to assess the advantage of considering a novel and sophisticated column generation approach with respect to a more straightforward MIP adaptation.  

\textbf{Computational environment:} Our numerical experiments have been conducted on a PC, with an Intel R CoreTM i7-1065G7 CPU @ 1.30GHz 1.50 GHz processor and 16 GB of RAM. The operating system is 64 bits. For the implementation of the column generation framework, we have adapted the column generation based estimators from \citet{krunalCG}, without applying the branch-and-price scheme (see above). We used Python 3.11 for all computations and Gurobi 9.11 \citep{Gurobi} for solving all LPs and MIPs and for integrating the classifiers, both the LR and the NNs.

\textbf{Experimental setup:} We calculate counterfactual explanations for two types of models: Neural Networks (NNs) and logistic regression (LR). We split our datasets randomly into train (50\%) and test sets (50\%), constructing the explanations only for the test instances. We also train three logistic regression models (one per dataset) with a regularization parameter $C=10$. Table \ref{tab:models} reports the test accuracy for each dataset, as well as the structure of the NN used.

\begin{table}[H]
    \centering    
    \caption{Model's Structure and Accuracy}\label{tab:models}
    \resizebox{0.9\columnwidth}{!}{%
    \begin{tabular}{lrrrr}
        \toprule
         & \multicolumn{3}{c}{\textbf{Neural Network}} & \multicolumn{1}{c}{\textbf{Logistic Regression}} \\
        \cmidrule(lr){2-4} \cmidrule(lr){5-5}
       \textbf{Dataset} & \textbf{Width} & \textbf{Depth} & \textbf{Accuracy} & \textbf{Accuracy}\\
        \midrule
        COMPAS  & 10  & 2 & 0.670 & 0.671 \\
        German Credit  & 20  & 2 & 0.728 & 0.724 \\
        Students performance & 14&3 &0.601 & 0.612\\
        \bottomrule
    \end{tabular}}
\end{table}

We calculate counterfactual explanations for different sizes of the set $S$, namely, $|S| \in \{10,20,50,100\}$, and several values for the maximum number of features to be perturbed. For the \textbf{COMPAS} dataset, we consider $T_{\text{max}} \in \{2,3\}$, for the \textbf{German credit} dataset, $T_{\text{max}} \in \{5,10,15\}$ and for the \textbf{Students performance} dataset, $T_{\text{max}} \in \{10,15,20\}$. 

\textbf{Evaluation}: To evaluate our column generation methodology, we report for each combination of $S$ size and $T_{\text{max}}$ the difference in performance of the CG algorithm and MIP \eqref{eq:milp_obj}-\eqref{eq:milp_cons8}. More precisely, we evaluate such difference in terms of the number of the counterfactual explanations that the two approaches can compute within a time limit of 3600 CPU seconds. For the CG formulation, we do not implement the branch-and-price scheme, thus, in principle, it might compute a suboptimal solution because of the lack of the necessary counterfactuals \emph{not} generated at the root node. In the vast majority of the cases, the value or ceiling of this turns out not to be the case\footnote{Precisely, in 94.1\% of the cases where the column generation does not reach the time limit, the final solution is certified to be optimal} but the gap reported in the evaluation can in principle be positive (the MIP solution being smaller), negative (the CG solution being smaller), or 0 when the two solutions have the same value. 

\subsection{Evaluation and analysis}

In Tables \ref{tab:compasLR}-\ref{tab:studentsNN}, the evaluation of the column generation framework is detailed. Specifically, for different values of the number of negative examples, $|S|$, and the maximum number of features allowed to change, $T_{\text{max}}$, we report the smallest number of explanations obtained, $y^{*}$, the gap between the CG and the MIP solutions, and the respective runtimes. 

We can see in Table \ref{tab:compasLR} for $|S|=10$ that for small values of $|S|$, when dealing with a dataset with few features like \textbf{COMPAS}, the MIP is faster. However, for higher values of $|S|$, for the rest of the datasets and especially with NN as ML model, using the column generation framework is not only considerably faster, but also, when using the MIP we are not able to reach the same quality of the solution within the time limit. Notice that in those cases, the gap is always\footnote{The only exception is in Table \ref{tab:compasLR} for $|S|=100$ and $T_{\text{max}}=3$, where the best solution obtained by the MIP is one unit smaller than the one of the CG algorithm.} negative, i.e., not only solving the MIP does not certify optimality but the number of explanations is higher than the value obtained with the CG methodology. For example, in Table \ref{tab:compasLR} for $|S|=100$ and $T_{\text{max}}=2$, a gap of -7 indicates that the solution of the MIP reached the time limit of 3600 CPU seconds with a set of 16 explanations, while the CG certifies that 9 are enough. 

Overall, the results in Tables \ref{tab:compasLR}-\ref{tab:studentsNN} clearly indicates that (i) the CG is an effective and scalable approach to compute small, structured counterfactual sets, and (ii) a naive approach based on the extension of a MIP framework from the literature is significantly outperformed by the CG one. 

The effect of the ML model used within the optimization framework (both CG and MIP) to classify the counterfactual explanations (see Section \ref{sec:classification}) is less clear from the experiments. 
\begin{itemize}
	\item On the one side, the differences of accuracy between NNs and LR are so minor that it is hard to define a trend: ideally, a more accurate model could require less counterfactuals and this intuition seems to be confirmed by a couple of results in the tables. Namely, for the \textbf{German credit} dataset for $|S|=20$ and $T_{\text{max}}=10$, we obtain 8 explanations for the LR whereas only 5 are needed by the NN that is slightly more accurate (see Table \ref{tab:models}). Conversely, for the \textbf{Students performance} dataset for $|S|=20$ and $T_{\text{max}}=10$, the slightly more accurate LR model needs 10 explanations versus the 11 of the NN. 
	\item On the other side, from an optimization standpoint, modeling NNs requires more constraints and variables, i.e., larger optimization models and Tables \ref{tab:compasLR}-\ref{tab:studentsNN} clearly show that the CG approach does not suffer from scalability issues as much as the MIP does. Thus, one can assume that for datasets where more advanced ML techniques (like NNs) are required to obtain satisfactory classification, CG would be the only option.
\end{itemize}
\begin{table}[h]
\begin{minipage}{0.46\textwidth}
    \centering
    \caption{Evaluation for the \textbf{COMPAS} dataset for the LR model for different values of $|S|$ and $T_{\text{max}}$}
    \resizebox{0.98\columnwidth}{!}{
    \begin{tabular}{rrrrrr}
        $|S|$&  $T_{max}$& $y^{*}$  &  Gap &CG time &MIP time  \\ \hline
        10 & 2 & 4 & 0 & 2.82 & \textbf{1.67} \\
        20 & 2  & 6  &  0  &   \textbf{6.25}   &     70.92   \\
        50 &  2 & 9  &  0  &    \textbf{106.36}  &     3600.00   \\
        100 & 2  & 9  &  \textbf{-7}  &   \textbf{728.43}   &    3600.00    \\
        10 &  3 &  2 &  0  &   2.08   &     \textbf{0.72}   \\
        20 & 3  &  3 & 0   &   \textbf{9.73}   &     16.57   \\
        50 &  3 & 4  &  0  &   1102.49   &    \textbf{193.4}    \\
        100 &  3 &  5 &  1  &   \textbf{3527.70}   &    3600.00    \\
    \end{tabular}}
    \label{tab:compasLR}
\end{minipage}\hfill%
\begin{minipage}{0.46\textwidth}
    \centering
    \caption{Evaluation for the \textbf{COMPAS} dataset for the NN for different values of $|S|$ and $T_{\text{max}}$}
    \resizebox{0.98\columnwidth}{!}{
    \begin{tabular}{rrrrrr}
        $|S|$&  $T_{max}$&$y^{*}$  &  Gap &CG time &MIP time  \\ \hline
        10 & 2 & 4 & 0 & \textbf{2.15} & 3600.00 \\
        20 & 2  & 6  &  0  &   \textbf{6.36}   &     3600.00  \\
        50 &  2 & 9  &  \textbf{-13} &   \textbf{92.02}  &     3600.00   \\
        100 & 2  & 9  & \textbf{-35}  &   \textbf{1285.20}   &  3600.00    \\
        10 &  3 &  2 &  0  &   3.99   &    \textbf{2.12}  \\
        20 & 3  &  3 & 0   &   \textbf{17.38} &     3600.00  \\
        50 &  3 & 4  &  0  &   \textbf{176.32}  &    3600.00   \\
        100 &  3 &  5 & 1  &   \textbf{3556.28}   &   3600.00\\
    \end{tabular}}
    \label{tab:compasNN}
\end{minipage}
\end{table}

\begin{table}[h]
\begin{minipage}{0.46\textwidth}
    \centering
    \caption{Evaluation for the \textbf{German credit} dataset for the LR model for different values of $|S|$ and $T_{\text{max}}$}
    \resizebox{0.98\columnwidth}{!}{
    \begin{tabular}{rrrrrr}
        $|S|$&  $T_{max}$& $y^{*}$  &  Gap &CG time &MIP time  \\ \hline
        10 & 5 & 9 & 0 & \textbf{5.31} & 548.38 \\
        20 & 5  &19  &  0  &  \textbf{21.62}   &     3600.00   \\
        50 &  5 & 38  &  \textbf{-12}  &   \textbf{1676.81}  &     3600.00   \\
        100 & 5  & 90  &  \textbf{-10}  &   3600.00   &  3600.00    \\
        10 &  10 & 4  &  0  &   \textbf{13.32}   &     109.08 \\
        20 & 10  &  8 & \textbf{-1}   &   \textbf{432.72} &  2874.25  \\
        50 &  10 &  15 &  \textbf{-35} &   3600.00  &    3600.00   \\
        100 &  10 & 58 &  \textbf{-42} &  3600.00   & 3600.00       \\
          10 &  15 & 2 &  0  &   18.55   &   \textbf{14.84}   \\
        20 & 15  & 4  &  0  &  \textbf{482.71}   &   3600.00    \\
        50 &  15&  6 &  \textbf{-1}  &   3600.00  &    3600.00   \\
        100 &  15 & 11 &  \textbf{-80} &  3600.00   &     3600.00  \\
    \end{tabular}}
    \label{tab:germanLR}
\end{minipage}\hfill%
\begin{minipage}{0.46\textwidth}
    \centering
    \caption{Evaluation for the \textbf{German credit} dataset for the NN for different values of $|S|$ and $T_{\text{max}}$}
    \resizebox{0.98\columnwidth}{!}{
    \begin{tabular}{rrrrrr}
        $|S|$&  $T_{max}$& $y^{*}$  &  Gap &CG time &MIP time  \\ \hline
        10 & 5 & 9 & 0 & \textbf{6.40} & 3600.00 \\
        20 & 5  & 19 & 0& \textbf{67.52}   &     3600.00   \\
        50 &  5 & 38  &  \textbf{-12}  &  \textbf{2139.81}    &   3600.00    \\
        100 & 5 & 90  &  \textbf{-10}  &   3600.00   &  3600.00    \\
        10 &  10 &  4 &  0  &   \textbf{23.49}  &     3600.00  \\
        20 & 10  &  5 & \textbf{-1}   &   \textbf{499.60}  &    3600.00  \\
        50 &  10 & 15  &  \textbf{-21}  &   3600.00  &    3600.00    \\
        100 &  10 &  64 & \textbf{-36}   &  3600.00   &   3600.00     \\
          10 &  15 & 2 &  0  &  \textbf{10.22}   &  79.35  \\
        20 &15  & 4  &  0  &  \textbf{459.90}   &    3600.00   \\
        50 &  15&  6 &  \textbf{-2}  &   3600.00  &   3600.00  \\
        100 &  15 & 11 & \textbf{-87}   &  3600.00   &    3600.00   \\
    \end{tabular}}
    \label{tab:germanNN}
\end{minipage}
\end{table}

\begin{table}[h]
\begin{minipage}{0.46\textwidth}
    \centering
    \caption{Evaluation for the \textbf{Students performance} dataset for the LR model for different values of $|S|$ and $T_{\text{max}}$}
    \resizebox{0.98\columnwidth}{!}{
    \begin{tabular}{rrrrrr}
        $|S|$&  $T_{max}$& $y^{*}$  &  Gap &CG time &MIP time  \\ \hline
        10 & 10 & 7 & 0 & \textbf{29.726} & 587.94 \\
        20 & 10  & 10  &  \textbf{-1}  &  \textbf{689.34}   &     3600.00   \\
        50 &  10 & 33  &  \textbf{-15}  &    3600.00  &     3600.00   \\
        100 & 10  & 77  &  \textbf{-23}  &   3600.00   &    3600.00  \\
        10 &  15 &  4 &  0  &   \textbf{34.20}   &     3071.42  \\
        20 & 15  &  5 & 0   &   \textbf{648.29} &    3600.00  \\
        50 &  15 & 12  &  \textbf{-6}  &   3600.00  &    3600.00    \\
        100 &  15 &  43 &  \textbf{-57}  &  3600.00   &    3600.00    \\
          10 &  20 & 2 &  0  &   \textbf{4.48}   &   5.85   \\
        20 & 20  & 2  &  0  &  87.41   &    \textbf{40.16}   \\
        50 &  20&  3 &  0  &   3600.00  &    3600.00   \\
        100 &  20 & 4 &  0  &  3600.00   &     3600.00  \\
    \end{tabular}}
    \label{tab:studentsLR}
\end{minipage}\hfill%
\begin{minipage}{0.46\textwidth}
    \centering
    \caption{Evaluation for the \textbf{Students performance} dataset for the NN for different values of $|S|$ and $T_{\text{max}}$}
    \resizebox{0.98\columnwidth}{!}{
    \begin{tabular}{rrrrrr}
        $|S|$&  $T_{max}$& $y^{*}$  &  Gap &CG time &MIP time  \\ \hline
        10 & 10 & 7 & 0 & \textbf{31.337} & 3600.00 \\
        20 & 10  & 11  &  0  &  \textbf{916.76}   &     3600.00   \\
        50 &  10 & 33  & \textbf{-15}   &    \textbf{3290.40}  &    3600.00   \\
        100 & 10  & 79  &  \textbf{-21}  &    3600.00  &   3600.00   \\
        10 &  15 &  4 &  0  &   \textbf{35.900}  &     3600.00  \\
        20 & 15  &  5 & \textbf{-1}   &   \textbf{930.42}  &    3600.00  \\
        50 &  15 & 12  &  \textbf{-36}  &   3600.00  &    3600.00    \\
        100 &  15 &  45 &  \textbf{-54}  &  3600.00   &    3600.00    \\
          10 &  20 & 2 &  0  &  \textbf{20.22}   &  48.21  \\
        20 & 20  & 2  &  0  &  \textbf{223.40}   &    2887.83   \\
        50 &  20&  3 &  \textbf{-46}  &   3600.00  &   3600.00  \\
       100 &  20 & 4 & \textbf{-92}  &  3600.00   &   3600.00   \\
    \end{tabular}}
    \label{tab:studentsNN}
\end{minipage}
\end{table}

\subsection{Drawing explanability}

\begin{figure}[h]
\centering
    \includegraphics[width=0.93\linewidth]{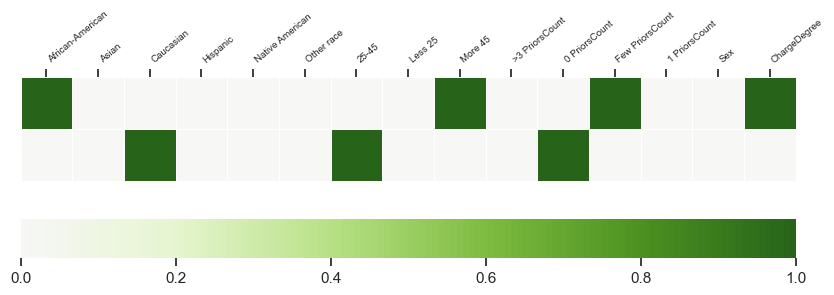}
    \caption{Explanations for \textbf{COMPAS} dataset for the NN, $|S|=10$ and $T_{\text{max}}=3.$}
    \label{fig:expl_compas2}
\end{figure}

\begin{figure}[h]
\centering
    \includegraphics[width=0.93\linewidth]{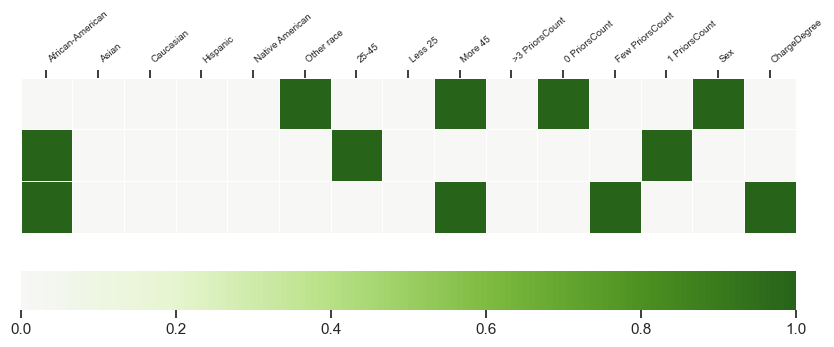}
    \caption{Explanations for \textbf{COMPAS} dataset for the NN,  $|S|=20$ and $T_{\text{max}}=3.$}
    \label{fig:expl_compas3}
\end{figure}

It is informative to picture the obtained counterfactuals, so as to make apparent that the explanability of the approach. For the \textbf{COMPAS} dataset, considering the NN and $T_{\text{max}}=3$, we  show the counterfactual instances for different sizes of $S$, namely, in Figure \ref{fig:expl_compas2}, we show the three explanations for the $|S|=10$. In the heatmap, each row is an explanation, and each column a feature. In green, we show the variables with a value of 1, and in white the ones that are 0. We visualize the same information for $|S|=20$ in Figure \ref{fig:expl_compas3}.

\begin{figure}[h]
\centering
    \includegraphics[width=0.95\linewidth]{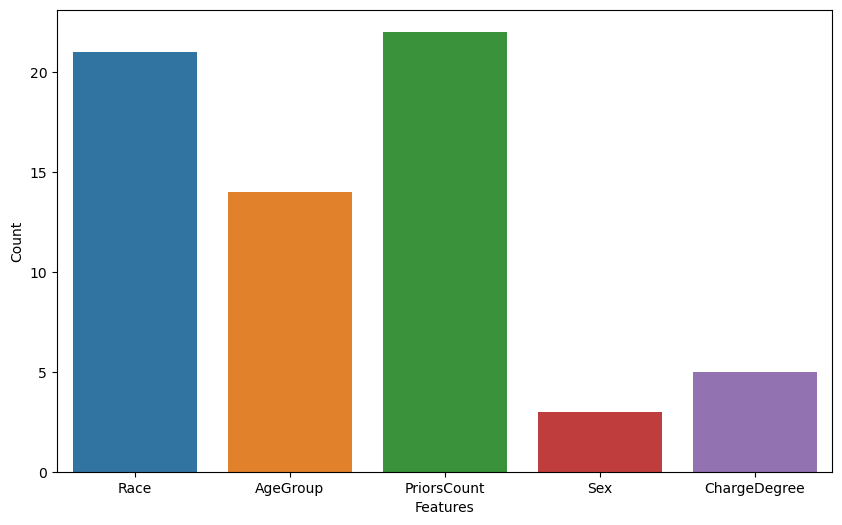}
    \caption{Number of times each of the features is perturbed for the \textbf{COMPAS} dataset for the NN}
    \label{fig:feat_compas}
\end{figure}

\begin{figure}[h]
\centering
    \includegraphics[width=0.95\linewidth]{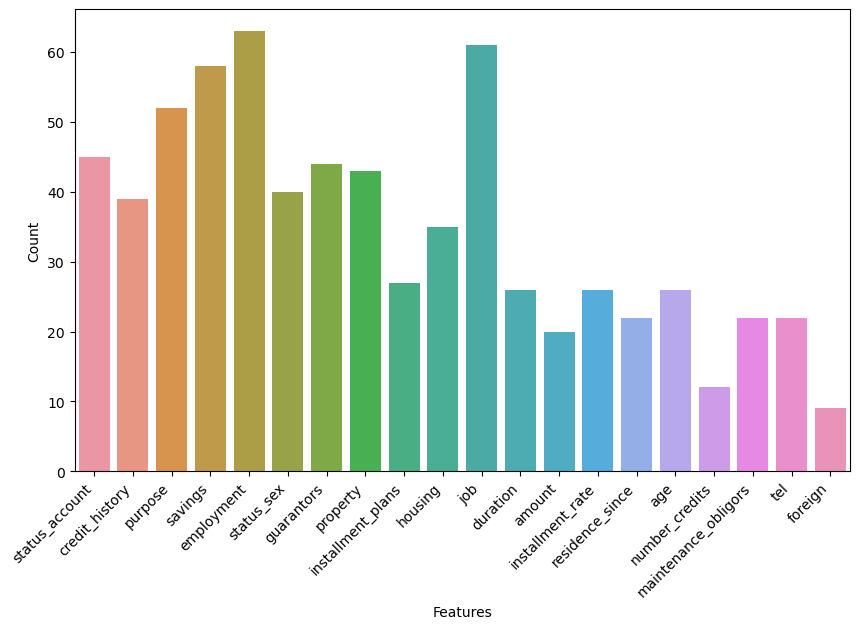}
    \caption{Number of times each of the features is perturbed for the \textbf{german credit} dataset for the NN}
    \label{fig:feat_german}
\end{figure}

\begin{figure}[h]
\centering
    \includegraphics[width=0.95\linewidth]{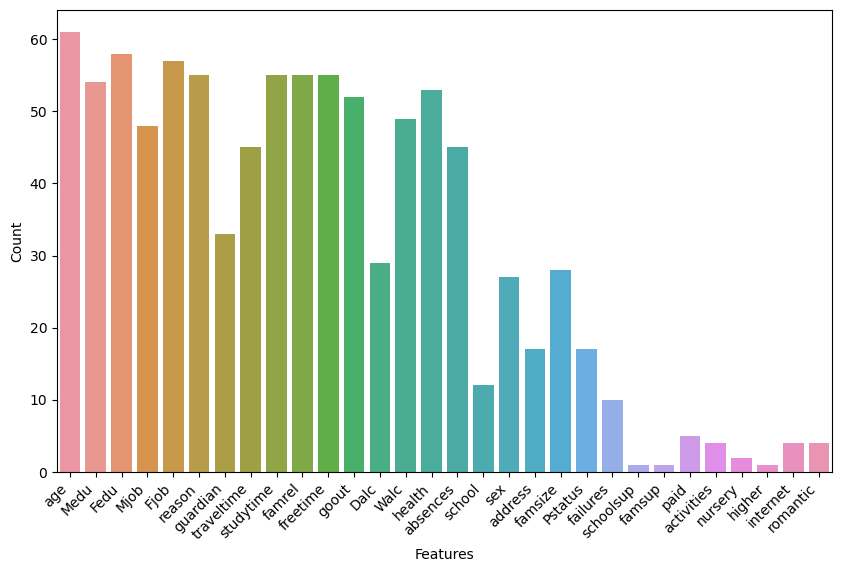}
    \caption{Number of times each of the features is perturbed for the \textbf{Students} dataset for the NN}
    \label{fig:feat_students}
\end{figure}

Finally, it is also worthwhile to see the most perturbed features. We can visualize for each of the datasets in the NN case, the number of total times one of the features is perturbed in Figures \ref{fig:feat_compas}-\ref{fig:feat_students}.

\section{Conclusions}
We considered the problem of generating a set of counterfactual explanations for a group of instances and, for the first time, we addressed explainability alltogerher (i) by minimizing the number of explanations and at the same time (ii) by considering sparsity in the number of features allowed to change. To do so, we proposed a novel column generation approach that has favorable properties in terms of scalability, computational performance and quality of the solutions.

\bibliography{bibliografia}


\end{document}